\pdfoutput=1

\documentclass[11pt]{article}

\usepackage{EMNLP2023}
\usepackage{tcolorbox}
\tcbuselibrary{skins,breakable}
\usepackage{times}
\usepackage{latexsym}
\usepackage{booktabs}
\usepackage{graphicx}
\usepackage{subcaption}
\usepackage{multirow}
\usepackage{soul}
\usepackage{amsmath}
\usepackage{float}
\usepackage[table]{xcolor}
\usepackage[T1]{fontenc}

\usepackage[utf8]{inputenc}

\usepackage{microtype}

\usepackage{inconsolata}

\usepackage{enumitem}
\title{MigrationNarrate: A Dataset for Detection of Migration Narratives in YouTube Videos}
\author{
Fatima Haouari, Carolina Scarton, Kalina Bontcheva  \\
Department of Computer Science, University of Sheffield, UK \\
\texttt{\{f.haouari,c.scarton,k.bontcheva\}@sheffield.ac.uk}
}
\begin{document}
\maketitle
\begin{abstract}
Narratives are central to how social communication is framed, making their detection critical for understanding and analysing public discourse. Prior work has explored narrative detection and extraction across diverse domains; however, migration narratives remain significantly understudied, primarily due to the absence of dedicated annotated datasets. Furthermore, public communication has recently shifted towards video-centric platforms, where narratives are conveyed through multimodal signals and consumed at scale. Despite this shift, narratives in videos remain largely unexplored. To bridge these gaps, we introduce \textbf{MigrationNarrate}, the first multimodal dataset for detection of migration narratives in the UK, consisting of 1,115 YouTube video transcripts annotated using a two-level taxonomy of 12 migration super-narratives and 53 narrative labels.  
This paper details the dataset design, collection, and annotations; together with benchmark results using a combination of pre-trained encoder models and both open- and closed-source Large Language Models. Finally, a thorough error analysis offers insights for future work.    

\end{abstract}

\section{Introduction}

Video has become a dominant form of content consumption. As for March 2026, YouTube reaches over 2.83 billion users worldwide, with approximately 340 million daily users and more than 1.8 million hours of videos uploaded daily~\cite{resourcera2026youtube}. This scale underscores the increasing influence of video-based media in shaping information exposure and public discourse, allowing narratives to be delivered at scale, potentially amplifying the spread of misleading or biased content. Despite a growing body of research in narratives detection and extraction, existing work has predominantly focused on textual data from social media platforms such X~\cite{ ai2024tweetintent, fraile2024automatic, haouari2025ukelectionnarratives} and Facebook~\cite{rowlands-etal-2024-predicting} and from news articles~\cite{coan2021computer, piskorski2022exploring, nikolaidis2025polynarrative}.
Detecting narratives in video transcripts remain largely understudied, despite its importance since videos differ from written text in structure and delivery~\cite{dingemanse-liesenfeld-2022-text}.

Additionally, although previous work has focused on diverse domains, such as COVID-19~\cite{kotseva2023trend, heinrich-etal-2024-automatic}, elections~\cite{haouari2025ukelectionnarratives}, climate change~\cite{coan2021computer, piskorski2022exploring, nikolaidis2025polynarrative}, and Ukraine-Russia war~\cite{ai2024tweetintent, nikolaidis2025polynarrative}, migration narratives are yet to be studied. 
\textit{Migration narratives} are defined by~\citet{JRC142039} as \textit``the stories, ideas and
perceptions that people have about migration and
migrants. These narratives can be influenced by various
factors such as media, politics and social interactions.
They can shape public opinion and policy decisions''.
While migration narratives can emphasise solidarity and humanitarian perspectives, migration-related discourse often contains polarised and negative framings that contribute to the spread of misconceptions and exclusionary attitudes~\cite{JRC142039}. This underscores the importance of detecting migration narratives for understanding and analysing public discourse.

\begin{figure}[t]
    \centering
    \includegraphics[width=0.95\columnwidth]{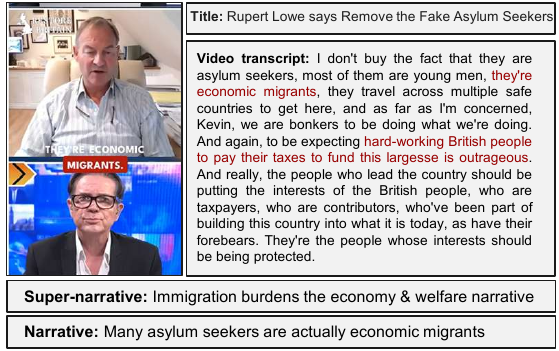}
    \caption{Example from \textbf{MigrationNarrate} dataset.}
    \label{fig:example}
\end{figure}

In this work, we take the first step towards addressing this gap by introducing \textbf{MigrationNarrate}, a novel dataset for migration narrative detection in videos (example in Figure~\ref{fig:example}). The dataset comprises 1,115 video transcripts annotated using a migration narrative taxonomy introduced by the European Commission’s Joint Research Centre (JRC)~\cite{JRC142039}. The taxonomy categorises European migration-related discourse into a set of recurrent narrative framings of \textbf{12 super-narratives and 53 narratives}, covering narratives that frame migration in humanitarian and solidarity-based terms, as well as narratives emphasising exclusion, policy regulation, or security concerns. 
To summarise, our contributions are as follows:

\begin{table*}[!htb]

\centering
\scriptsize
\begin{tabular}{l c c c c c}
\toprule
{\bf Dataset} & {\bf Source} & {\bf Topic} & {\bf Lang} & {\bf \# Narratives} & {\bf \# Instances} \\
\midrule
\bf CARDS~\cite{coan2021computer} 
& Articles 
& Climate Change 
& En 
& 5/27 
& 28,945 \\

\bf Climate~\cite{piskorski2022exploring} 
& Articles 
& Climate Change 
& En 
& 5/27 
& 1,118 \\

\bf COVID-19~\cite{kotseva2023trend} 
& Articles 
& COVID-19 
& En 
& 12/51 
& 58,625 \\

\bf COVID-19-Ger~\cite{heinrich-etal-2024-automatic} 
& Telegram 
& COVID-19 
& De 
& 14 
& 1,099 \\

\bf DIPROMATS~\cite{fraile2024automatic} 
& Twitter 
& Narratives around COVID-19 by Authorities 
& En/Es 
& 4/24 
& 1,272,272 \\

\bf TweetIntent@Crisis~\cite{ai2024tweetintent} 
& Twitter 
& Ukraine--Russia War
& En 
& 2 
& 3,691 \\

\bf Climate\_ads~\cite{rowlands-etal-2024-predicting} 
& Facebook 
& Climate Change in US ads 
& En 
& 7 
& 1,330 \\

\textbf{PolyNarrative~\cite{nikolaidis2025polynarrative}}& Articles 
& Ukraine--Russia War//Climate Change
& Multi
&11/38//10/36
& 2,467 \\
\textbf{UKElectionNarratives~\cite{haouari2025ukelectionnarratives}} 
& Twitter 
& UK Elections 2019/2024 
& En 
& 10/32 
& 2,000 \\
\midrule
\bf MigrationNarrate 
&  YouTube &  Migration Crisis in UK & En & 12/51 & 1,115 \\
\bottomrule
\end{tabular}

\caption{Comparison of \textbf{MigrationNarrate} with existing public datasets for detecting narratives. \# Narratives is presented as \# Super-narratives/\# Narratives for datasets with two-level annotation scheme.}
\label{tab:datasets}
\end{table*}

\begin{itemize}[leftmargin=*=*, nolistsep]
    \item \textbf{MigrationNarrate}, the first dataset for detecting UK-related migration narratives in video transcripts, consisting of 1,115 YouTube video transcripts annotated using a two-level taxonomy of migration super-narratives and narrative labels. 4,425 automatically filtered, unlabelled migration-related videos are also made available.
    \item Benchmarking pre-trained encoder models and both open- and closed-source Large Language Models (LLMs) on \textbf{MigrationNarrate}.
    \item An extensive error analysis providing insights for future research.
    
\end{itemize}

\section{Related Work}

\paragraph{Migration Narratives} Migration narratives have been widely studied in political and media contexts, both conceptually and empirically~\cite{sahin2020migration, maneri2023comparative}. Prior work defines their roles in policymaking and public discourse~\cite{dennison2021narratives}, while other studies examine their distribution across media platforms and countries, highlighting the influence of context, event type, and media genre~\cite{maneri2023comparative}. Additional work explores the factors driving narrative influence and success~\cite{dennison2021narratives, garces2025making}, as well as their circulation between media and political actors~\cite{smellie2025influence}. Policy-oriented research further provides a comprehensive overview of migration narratives and their impact on public perception and policy, defining migration narratives and introducing a hierarchical taxonomy of super-narratives and narrative categories~\cite{JRC142039}. Recently, \citet{ghafouri2025framing} analysed migration discourse in UK parliamentary debates and US congressional speech using open-source LLMs for stance annotation and narrative frame extraction, enabling the analysis of migration framing across time and political parties. However, they did not release a publicly annotated dataset for the task.
\paragraph{Narratives Detection}
As summarised in Table~\ref{tab:datasets}, prior work on narrative detection has examined a range of topics, including climate change~\cite{coan2021computer, piskorski2022exploring, rowlands-etal-2024-predicting, nikolaidis2025polynarrative}, the Ukraine war~\cite{ai2024tweetintent, nikolaidis2025polynarrative}, COVID-19~\cite{kotseva2023trend, heinrich-etal-2024-automatic, fraile2024automatic}, and elections~\cite{haouari2025ukelectionnarratives}. These studies analyse narratives across different platforms, primarily using social media such X~\cite{ai2024tweetintent, haouari2025ukelectionnarratives}, and Facebook~\cite{rowlands-etal-2024-predicting}, and news articles~\cite{coan2021computer, nikolaidis2025polynarrative} as data sources. However, to the best of our knowledge, there is no dataset specifically designed for the detection of migration narratives in video transcripts. 
To address this gap, we introduce \textbf{MigrationNarrate} the first dataset of migration narratives in video transcripts.

\section{The MigrationNarrate Dataset}
In this section, we present the migration narratives taxonomy adopted in our study (Sec~\ref{taxonomy}), our approach to collecting videos (Sec~\ref{videos_collection}), the process of filtering migration-related videos (Sec~\ref{videos_filtering}), and the annotation procedure (Sec~\ref{annotation}).

\subsection{Migration Narratives Taxonomy}\label{taxonomy}
In this work, we adopt the migration narrative taxonomy introduced by the JRC~\cite{JRC142039}. This taxonomy categorises European migration-related discourse into a set of recurrent narrative framings identified through an extensive review of media coverage, policy debates, and prior literature. It comprises \textbf{12 super-narratives and 53 narratives}, covering narratives that frame migration in humanitarian and solidarity-based terms, as well as narratives emphasising exclusion, policy regulation, or security concerns. By covering multiple forms of migration framing, the JRC taxonomy is well suited for labelling and analysing migration narratives in our data. 

The JRC taxonomy provides explicit definitions for super-narratives, which define the broader conceptual scope of their associated narratives. Although it does not provide direct definitions for individual narratives, their meanings can be inferred from the super-narratives. To support annotation, we prompted GPT-5.1~\cite{singh2025openai} to generate concise narrative definitions based on their corresponding super-narrative definitions, which were subsequently manually reviewed and validated by the authors to ensure narrative and super-narrative alignment (see  Appendix~\ref{sec:narratives_def} for the full set of narrative definitions). 




\subsection{YouTube Videos Collection}\label{videos_collection}
To collect videos \textit{potentially} discussing the migration crisis in the UK, we adopted two approaches:
\paragraph{Channel- and Playlist-Based Video Collection}
We collected YouTube videos using the YouTube Data API v3,\footnote{\url{https://developers.google.com/youtube/v3}}
 targeting both channels whose content is primarily centred on migration discourse and channels with a broader editorial scope where migration-related discussions frequently appear.
We selected channels to reflect influential sources in the UK from multiple perspectives, including institutional media outlets (e.g., BBC News and GB News), political actors such as political parties and government channels (e.g., Labour Party and Reform UK), as well as prominent non-institutional commentators, activists, and independent accounts that regularly engage in migration-related debates (e.g., @TommyRobinsonOnline and @drmatthewgoodwin).
For channels focused on political actors or organisations, such as political parties we collected all available videos uploaded in their channels. For channels with more heterogeneous content, such as news agencies, we collected videos selectively by targeting specific playlists. Playlist selection included both explicitly migration-focused playlists (e.g., \textit{Migration Crisis} on Sky News and \textit{Immigration} on TalkTV) and broader political playlists where migration-related content may also appear (e.g., \textit{UK Politics} on Channel 4 News).
Overall, videos were collected from 25 playlists and 21 unique channels. 
The full list of channels and playlists is presented in Appendix~\ref{appendix:videos_collection}.


\paragraph{Search-Based Video Collection} 
In addition to collecting videos from channels and playlists, we employed a complementary search-based retrieval strategy to capture migration-related content produced by a broader and more diverse set of users. This approach aimed to identify videos containing migration narratives beyond institutional, political, or media-affiliated sources. We used the YouTube Search API\footnote{\url{https://developers.google.com/youtube/v3/docs/search/list}} to retrieve videos using a set of general and narrative-oriented search phrases. The initial set of \textbf{general search phrases} was constructed based on domain knowledge and commonly used terms in public discussions of migration in the UK. To improve coverage and diversify the query space, \textbf{narrative-oriented search phrases} were constructed to retrieve videos that potentially express these narratives, and subsequently refined through multiple rounds of manual review.
In total, \textbf{77 search phrases} were used (listed in Appendix~\ref{search_phrases}) including \textbf{15 general phrases} such as \textit{Channel crossings UK}, and \textit{Stop the boats UK}, as well as \textbf{62 narrative-oriented phrases} such as \textit{immigration housing crisis UK} and \textit{immigrants taking jobs UK}, which corresponds to the \textit{Immigration burdens the economy \& welfare} super-narrative. 

\subsection{Filtering Migration videos}\label{videos_filtering}
Given that annotation is costly and time-consuming, particularly in the context of large-scale and diverse migration narratives, we employ a multi-step filtering pipeline. By removing videos that are unlikely to contain migration narratives, cost and annotators time / effort are significantly reduced.
\paragraph{Metadata-based filtering} We downloaded the collected videos using \textit{yt-dlp}, an open-source tool for video downloading from YouTube.\footnote{\url{https://github.com/yt-dlp/yt-dlp}}
        Only videos with a duration of \textit{three minutes or less} and published between \textit{1 January 2024 and 30 September 2025} were retained, focusing the dataset on short-form content. Following downloading the videos, we generated the video transcripts using \textit{faster-whisper}~\cite{radford2023robust}, an efficient implementation of the Whisper automatic speech recognition model.\footnote{\url{https://github.com/SYSTRAN/faster-whisper}} 
        Videos without successfully generated transcripts were excluded. 
        In addition, a subset of videos retrieved via the YouTube Search API (particularly those published in 2024) were no longer accessible at the time of download, despite having retrievable metadata. These cases corresponded to videos that had been removed or whose channels were no longer available. This filtering step resulted in a total of 20,992 videos.

    \paragraph{Semantic filtering} Aim to identify UK migration-relevant content, we filtered video transcripts using embedding-based semantic similarity. Migration relevance was estimated using the E5-Large model~\cite{wang2022text}\footnote{\url{https://huggingface.co/intfloat/e5-large-v2}} by comparing transcript embeddings against a fixed set of ten immigration-related queries (queries are presented in Appendix~\ref{appendix:queries}). For each transcript, cosine similarity was computed with each query, and the maximum similarity score was taken as the immigration score.
    The query set was designed to capture broad and commonly occurring formulations of migration-related discourse in the UK context. Queries were expressed using general, high-level language rather than specific slogans or claims, with examples including \textit{``immigration in the UK''} and \textit{``asylum seekers in the UK''}. Transcripts with an immigration score of 0.8 or higher were retained for subsequent analysis.\footnote{This threshold was selected based on a manual inspection of a representative sample, balancing coverage and precision.} In total 5,540 videos were selected at this stage.
   
    \paragraph{A diverse sample for human annotation} 
    In order to pre-select a balanced subset of videos across super-narratives and narratives for subsequent human annotation~\cite{ai2024tweetintent, haouari2025ukelectionnarratives},  we used GPT-5.1 for an initial round of pre-annotation. Given a video transcript and the migration narrative taxonomy, the model was prompted to identify the dominant super-narrative and narrative expressed in the video. It was also instructed to return a confidence score for its prediction, with \textit{None} assigned when no migration narrative was explicitly stated or strongly implied.
    Videos with a main narrative confidence score of at least 0.8 were selected, and videos assigned the \textit{None} label were excluded. For each narrative category, if the number of eligible videos was 40 or fewer, all such videos were retained. When a narrative contained more than 40 eligible videos, 40 videos were sampled using a round-robin strategy designed to promote diversity across publication year (2024 vs. 2025) and channel title, ensuring balanced temporal coverage and limiting over-representation from individual channels. A total of \textbf{1,115 videos} were selected for human annotation.

\subsection{Annotation Management}\label{annotation}
Given the complexity of the annotation task, we did not rely on crowdsourcing; instead, we recruited 6 students majoring international relations and politics (undergraduate in their final year and postgraduate students), all of whom where native or highly proficient in English. We compensated the annotators at a rate of 17.58 GPB per hour, and we conducted the annotation using the collaborative web-based annotation tool Teamware~\cite{wilby-etal-2023-gate}.\footnote{\url{https://gatenlp.github.io/gate-teamware/development/}} The annotation process was as follows:

\paragraph{Pilot study} Before commencing the official annotation, we conducted a pilot study with two recruited students, both with prior experience in annotating political narratives. The pilot study helped identify key challenges annotators might face and informed refinements to the annotation guidelines, task setup, and training. This phase also resulted in a gold-standard set of 30 videos with full annotator agreement, which was further verified by the authors and used for quality control.
\paragraph{Annotators training}
Following the pilot study, all potential annotators were invited to a training session to introduce them to the project. This session covered the annotation guidelines, included illustrative examples, and provided an opportunity for questions and discussion to ensure a shared understanding of the task. After training, annotators labelled the gold set, and only those who achieved a Cohen' kappa of at least 0.45 on super-narratives and 0.35 on narratives were selected for the main annotation phase. These thresholds were selected based on our pilot study, as well as prior work showing only fair to moderate agreement are achieved during the early stages of annotating fine-grained labels~\cite{salman-etal-2023-detecting, haouari2025ukelectionnarratives}.

\paragraph{Data annotation} We adopted a batch-based annotation scheme~\cite{maarouf-etal-2024-hqp, haouari2025ukelectionnarratives}, partitioning the data into 11 batches. Each batch was independently annotated by two annotators and one adjudicator who resolves disagreements and determines the final label, following established practices in narrative annotation~\cite{nikolaidis2025polynarrative, piskorski-etal-2025-semeval}. Annotators were provided with the video transcript and a direct link to the video (see Figure~\ref{annotation_framework} in Appendix~\ref{appendix:annotation_framework}). To reduce cognitive load caused by the large number of narrative labels (53 labels), annotators were initially shown only the list of super-narratives (12 labels), while the full list of individual narratives was hidden. Once a super-narrative was selected, the corresponding set of narratives was displayed, allowing annotators to choose the most appropriate label while reducing confusion and improving annotation consistency. Definitions of all super-narratives and individual narratives were provided to annotators. During annotation, these definitions were accessible by hovering over each label on the annotation framework, allowing annotators to quickly review the intended meaning of a narrative. Additionally, a separate document containing the full set of definitions was shared during training so that annotators could review the taxonomy before the annotation session.

\paragraph{Data quality}Inter-annotator agreement measured using Cohen's kappa before resolving disagreement was 0.45 at super-narratives and 0.30 at narratives level showing moderate and fair agreement respectively. These agreement levels are modest but expected given the task’s complexity and fine-grained labels~\cite{nikolaidis2025polynarrative, piskorski-etal-2025-semeval, haouari2025ukelectionnarratives}. To resolve disagreement cases, the dataset was further refined through a final consolidation step conducted by an adjudicator, selected from the most experienced annotators or one of the authors with expertise in similar political narratives annotation tasks.






\subsection{Data Statistics}
The resulting \textbf{MigrationNarrate} dataset consists of 1,115 human-annotated videos collected from 684 unique YouTube channels, as summarised in Table~\ref{tab:dataset_size} (Appendix~\ref{stats}). These videos represent a manually selected and carefully annotated subset of a substantially larger corpus that was automatically filtered using our multi-step filtering pipeline presented in Section~\ref{videos_filtering}.
In addition to the human-annotated set, \textbf{we release the larger filtered corpus} which can support weak and semi-supervised learning, enabling scalable narrative classification beyond fully supervised settings~\cite{chen2024open, yang2025calibrating}. Moreover, the filtered corpus can be manually annotated to expand our dataset.

Table~\ref{tab:statistics} presents the distribution of super-narratives and individual narratives in \textbf{MigrationNarrate}. The most prevalent super-narrative is \textit{Immigration burdens the economy \& welfare narrative}, with 190 videos, while the most frequent individual narrative is \textit{Political rivals’ policy failed/ is failing/ will fail} (under the \textit{It is Us vs. Them (the establishment)} super-narrative), with 65 videos.
Additionally, similar to existing datasets with fine-grained narrative labels~\cite{haouari2025ukelectionnarratives, nikolaidis2025polynarrative, piskorski-etal-2025-semeval}, the class distribution is imbalanced. Table~\ref{tab:domain_examples} in Appendix~\ref{examples_appendix} shows examples from  \textbf{MigrationNarrate}.

\begin{table}[!hbt]
\tiny

\begin{tabular}{p{6.7cm}c}
    \toprule
     \textbf{Immigration burdens the economy \& welfare narrative} &\bf190  \\
            Immigrants do not contribute to the economy &  25\\
          Immigrants take our jobs& 10   \\
       Many asylum seekers are actually economic migrants&21
        \\
        Natives loose to immigrants/Immigrants receive better benefits than nationals &  32\\
          Receiving immigrants is too expensive/immigration burdens tax payers&32\\
        Immigrants abuse the welfare system& 16   \\
      
          Immigrants are a strain on the housing market &33  \\
           Immigrants are a strain on our health-care system&5
        \\
      
       Immigration reduces the attractiveness of location and lowers overall life quality&6\\
     
        Natives first& 10   \\
         \midrule


       \textbf{It is Us vs. Them (the establishment)} &\bf177  \\
           Political rivals’ policy failed/ is failing/ will fail&65\\
        Political rivals act against the interests of the people &39  \\
            Discrediting liberal values &  2\\
          Discrediting political rivals (e.g. ridicule, accusation)&42    \\
        
       Let's overthrow the establishment& 7   \\
   
      EU is harmful for our nation&4\\
         Our sovereignty is under threat&1\\
      Anti-Elitism& 17   \\
       
     \midrule


       \textbf{Immigrants are victims} &\bf115  \\
        Immigrants are victims of false hope& 12\\
        Immigrants are victims of human traffickers&30   \\
        EU/MS do not have the means to properly take care of immigrants&3\\
        Certain immigrants/asylum seekers are discriminated against&34 \\
       MS or EU are in breach of international law&  12  \\
     Immigrants suffer from labour exploitation&24\\

     \midrule


        \textbf{Immigration is a threat} & \bf111 \\
         Immigration is a threat to individual safety& 26 \\
       Immigration is a threat to national security& 21 \\
        Immigrants are prone to committing crimes (violent, non-violent, or organised)&  22\\
        Immigrants are prone to committing crimes of a sexual nature &29\\  Many immigrants are terrorists&13\\

    \midrule

      \textbf{Immigrants’ identity and/or culture is problematic} &  \bf 104\\
            Immigration is a threat to the European way of life/identity & 27 \\
          Arabs and/or Muslims are a cultural/social threat&38    \\
       Certain immigrants are unwilling/incapable to integrate&39
        \\
     \midrule
       \textbf{We are competent} & \bf90 \\
            Promoting policy initiatives or ideas &53\\
        Applauding political allies & 4   \\
        We represent the people &7\\
        We stand for common sense &6  \\
       We will reinstate order &20  \\

          \midrule
        \textbf{Immigration is out of control} & \bf 82\\
        There are too many immigrants coming & 48 \\
        Our borders are strained and/or insufficiently controlled  &34   \\
    
    \midrule
       \textbf{There is an unfair bias against our political camp} & \bf 43\\
         We are being silenced & 25 \\
        Mainstream media are biased& 5 \\
        There are double-standards/the system is corrupt&11\\
        The judicial system favours political rivals and/or obstructs own political camp &2 \\
            \midrule
      \textbf{Pragmatic approaches to immigration} & \bf43 \\
         Immigration helps tackle social \& economic issues & 43 \\
       
        \midrule
        \textbf{Immigration is part of a conspiracy} & \bf33  \\
        Blaming global elites &11  \\
        Great replacement & 17 \\
        Other conspiracy theory & 5 \\
    \midrule
    \textbf{Narrative strategy to remain in the news} &\bf 11  \\
         Reminder that immigration is an issue & 5 \\
        Self-promotion &6 \\

    \midrule
       
        \textbf{Geopolitical narratives: Countries are acting against the interest of other countries} &\bf 10\\
        Immigration is abused as a political tool &2  \\
        Our country is doing more than other countries to deal with immigration issues &  8 \\
        
    \midrule

     \textbf{None} & 106 \\
    \midrule
    \textbf{Total} &\bf1,115\\
    \bottomrule
\end{tabular}%
\caption{\textbf{MigrationNarrate} super-narratives (in bold) and narratives statistics.  }
\label{tab:statistics}
\end{table}

\section{Experiments and Evaluation}
To establish baselines for the \textbf{MigrationNarrate} dataset, 
we benchmarked different model configurations across three dimensions: \textit{(1) pre-trained encoder-based models versus LLMs}, \textit{(2) prompting-based inference versus supervised fine-tuning}, and \textit{(3) open-source versus closed-source models}. 
Our experiments rely exclusively on video transcripts, providing a strong text-based baseline for the task. More advanced approaches that incorporate additional modalities, such as visual and audio signals from the videos, remain an important direction for future work.
\begin{table*}[!htb]
\small
\centering
\renewcommand{\arraystretch}{1.2}
\setlength{\tabcolsep}{4pt}
\scalebox{0.86}{
\begin{tabular}{l l ccc ccc}
\toprule
\multirow{2}{*}{\textbf{Model}} 
& \multirow{2}{*}{\textbf{Setting}}
& \multicolumn{3}{c}{\textbf{Super-narratives}} 
& \multicolumn{3}{c}{\textbf{Narratives}} \\
\cmidrule(lr){3-5} \cmidrule(lr){6-8}
 &  & \textbf{Macro-F1} & \textbf{Macro-Prec} & \textbf{Macro-Recall}
    & \textbf{Macro-F1} & \textbf{Macro-Prec} & \textbf{Macro-Recall} \\
\midrule

Roberta-large
  & fine-tuned  & 0.446 & 0.445 & 0.458& 0.212& 0.223 &0.234 \\
\hline
\multirow{3}{*}{Llama-3.1-8B-Instruct}
  & zero-shot  & 0.226 & 0.307 & 0.224& 0.130& 0.203 & 0.132\\
  & few-shot   & 0.115 & 0.296 & 0.132& 0.028 & 0.055 & 0.041 \\

  & fine-tuned & \bf0.525& 0.543 & 0.525& 0.304 & 0.309& 0.340 \\
\hline
\multirow{3}{*}{Qwen2.5-7B-Instruct}
  & zero-shot  & 0.246& 0.315 & 0.272&0.107 & 0.137 & 0.114\\
  & few-shot   & 0.316 & 0.350 & 0.329& 0.160 & 0.188 &0.182\\
  & fine-tuned & 0.434 & 0.472 & 0.449& 0.249 &0.261 & 0.278 \\
\hline
\multirow{3}{*}{Gemma-3-12b-it}
  & zero-shot  & 0.378 & \underline{0.552} & 0.395& 0.227& 0.278 & 0.254 \\
  & few-shot   &0.423& 0.414& 0.467&0.228 & 0.235& 0.267 \\
  & fine-tuned & \underline{0.502} & 0.512 & \bf0.521& 0.282 & 0.287&0.304 \\

\hline
\multirow{2}{*}{GPT-4o}
  & zero-shot  & 0.427 & 0.500 & 0.455 & 0.305 & \bf0.346 & 0.353\\
  & few-shot   & 0.440 & 0.460 & \underline{0.509}& \bf0.350& \underline{0.341} & \bf 0.403 \\
\hline
\multirow{2}{*}{GPT-5.4}
  & zero-shot  &0.466 & \bf 0.558 & 0.472& 0.316 & \underline{0.341} & \underline{0.380} \\
  & few-shot   & 0.480 & 0.503 & 0.502& \underline{0.324} & 0.323 & 0.373\\
  \bottomrule
\end{tabular}
}
\caption{Performance of the models on the \textbf{test set}. The best score for each evaluation measure is shown in bold, and the second-best score is underlined. }
\label{tab:model_results}
\end{table*}
\subsection{Experimental setup}

\paragraph{Models Selection} we experimented with the following set of models:
\begin{enumerate}[leftmargin=*=*, nolistsep]
\item \textbf{Pre-trained encoder-based models}: we fine-tuned RoBERTa-large~\cite{liu2019roberta} for two classification tasks: (1) a model that predicts the higher-level super-narratives; (2) a model that predicts the fine-grained narratives.
\item \textbf{Open-source LLMs}: we evaluated Llama-3.1-8B-Instruct~\cite{grattafiori2024llama3herdmodels}, Qwen2.5-7B-Instruct~\cite{qwen2025qwen25technicalreport}, and Gemma-3-12b-it~\cite{gemmateam2025gemma3technicalreport} under \textit{zero-shot}, \textit{few-shot}, and \textit{fine-tuning} setup.

\item \textbf{Closed-source LLMs}: we prompted GPT-4o~\cite{hurst2024gpt} and GPT-5.4~\cite{openai2026gpt54} under \textit{zero-shot} and \textit{few-shot} setup.
\end{enumerate}
\paragraph{Prompting setup}
For all LLMs, we adopted a two-step hierarchical prompting approach~\cite{singh2025gatenlp} in which the models were first asked to identify the super-narrative and then to detect the corresponding narrative. This setup mirrors the annotation procedure provided to annotators and helps avoid confusing the models with a long list of narratives. 
For both the \textit{zero-shot} and \textit{few-shot} settings, the label definitions were provided to the models. In the \textit{few-shot} setting, the models were additionally given one example for each class. Our Prompts are presented in Appendix~\ref{appendix:benchmark_prompt}.

\paragraph{Fine-tuning setup}
We apply standard supervised fine-tuning to RoBERTa-Large, while the three LLMs are adapted using Quantized Low-Rank Adaptation (QLoRA) \citep{dettmers2023qlora}. Detailed training settings and hyperparameter configurations for all models are in Appendix~\ref{sec:Hyperparameters} and \ref{gpu}.

\paragraph{Data splits}
To create the data splits, we used stratified sampling~\citep{sechidis2011stratification}, allocating 70\% of the data for training, 10\% for development, and 20\% for testing. Specifically, 774, 117 and 224 videos as train, dev, and test  data respectively. For extremely low-frequency narrative classes containing only one or two instances (four narratives in total), these examples were excluded from the training split and retained only for evaluation.

\subsection{Results and Discussion}

\paragraph{Pre-trained encoders vs. LLMs:} As shown in Table~\ref{tab:model_results}, for \textbf{super-narrative classification} the fine-tuned RoBERTa model achieved results close to GPT-4o, slightly surpassing it in the \textit{few-shot} setting, and outperformed all open-source LLMs under both \textit{zero-shot} and \textit{few-shot} settings. However, it remained below GPT-5.4, which achieved a Macro-F1 of 0.480 compared to 0.446 for RoBERTa. Fine-tuning substantially improved the performance of 
Gemma and Llama, which surpassed RoBERTa with Macro-F1 scores of 0.502 and 0.525, respectively. However, the fine-tuned Qwen remained below the encoder-based baseline. At the \textbf{narrative level}, both Qwen and Llama underperformed compared to RoBERTa in the \textit{zero-shot} and \textit{few-shot} settings, whereas Gemma and GPT models outperformed RoBERTa. After \textit{fine-tuning}, all open-source LLMs consistently outperformed RoBERTa for fine-grained narrative detection, demonstrating that LLMs benefit substantially more from task-specific adaptation than from prompting alone for this task. 

\paragraph{Prompting vs. fine-tuning LLMs:} Prompting with \textit{few-shot} examples consistently improved performance over the \textit{zero-shot} setting for both super-narrative and narrative classification across almost all open- and closed-source models. The only exception was Llama, whose performance slightly declined when examples were introduced, suggesting that it may be more sensitive to prompt formulation or example selection. Finally, fine-tuning all open-source LLMs led to substantial performance gains compared to prompting alone, confirming that task-specific adaptation is significantly more effective than \textit{zero-shot} or \textit{few-shot} prompting for this task. 

\paragraph{Open- vs. closed-source:} For \textbf{super-narrative classification}, both GPT-4o and GPT-5.4 outperformed all open-source LLMs in \textit{zero-shot} and \textit{few-shot} settings. Notably, even in the \textit{zero-shot} setting, the closed-source models achieved higher macro-F1 than open-source LLMs with \textit{few-shot} prompt, highlighting their superior instruction-following capabilities. However, after fine-tuning, Llama achieved the best overall performance for \textbf{super-narrative classification}, with a Macro-F1 of 0.525 compared to the best results for GPT-4o (0.440) and GPT-5.4 (0.480), while Gemma also outperformed the closed-source models achieving a Macro-F1 of 0.502. For \textbf{narrative classification}, however, even fine-tuned open-source LLMs were unable to surpass the closed-source models. GPT-4o and GPT-5.4 achieved the best results, with Macro-F1 score of 0.350 and 0.324, respectively, while the two strongest open-source models, Llama and Gemma, reached Macro-F1 scores of 0.304 and 0.282, respectively. These findings suggests that fine-grained narrative detection remains particularly challenging for open-source models, even after task-specific fine-tuning.

\begin{table*}[!htb]
\centering
\renewcommand{\arraystretch}{1.35}
\setlength{\tabcolsep}{5pt}
\tiny

\begin{tabular}{p{2.8cm}|p{8.2cm}|p{1.7cm}|p{1.9cm}}
\hline

\textbf{Error Type / Explanation} 
& \textbf{Sentences from the video transcript} 
& \textbf{Gold Label} 
& \textbf{Model Predictions} \\
\hline
\textbf{Surface-Narrative Misalignment (49.7\%).}
The model predicts a narrative aligned with salient topical cues rather than the transcript’s dominant narrative, failing to identify the speaker’s primary argumentative focus.

&
``On Saturday this week, \textcolor{red}{a record 973 illegal migrants entered Britain on 17 boats. That is the biggest number this year.} And it's yet more evidence, as \textbf{I've long been arguing, for why Labour's plan for smashing the gangs isn't going to work. With no active deterrent, with no Rwanda-style plan, with no serious approach to dissuading people from crossing the Channel, with a massive backlog in the asylum system and a pro-immigration government, this crisis on the border is going to get worse and worse and worse.}''

&

Political rivals’ policy failed/ is failing/ will fail

&

\textbf{G:} There are too many immigrants coming \newline
\textbf{L:} Our borders are strained and/or insufficiently controlled
\\
\hline
\textbf{Fine-Grained Label Confusion (26.67\%).} The model predicts a semantically related but incorrect narrative label, failing to distinguish between closely related categories and misidentifying the dominant narrative.

&
\textbf{``British citizens are working hard, paying taxes and yet they can't even afford a home.} \textcolor{red}{But the same basically is given free to the illegal migrants who are entering into the country from the day one.} \textcolor{red}{I mean they are given forced-fire hotel facilities, NHS facilities, dentist facilities, refugees loans and what not.} \textbf{All paid by the basically hardworking taxpayer and it has not changed basically from the last successive governments. The situation is getting actually from bad to worse. And there is also basically a growing sense of frustration among the people who feels that the government is actually ignoring them. I think government really need to think about it that the hardworking people paying taxes can't afford the facilities and they have been given free to the people who are coming from other countries without contributing anything.}''

&

Receiving immigrants is too expensive/ immigration burdens tax payers

&

\textbf{G:} Natives loose to immigrants / Immigrants receive better benefits than nationals \newline
\textbf{L:} Natives loose to immigrants / Immigrants receive better benefits than nationals
\\
\hline
\textbf{Multi-Narrative Overlap (12.28\%).} The transcript contains multiple strong narratives, making more than one label plausible depending on which is treated as dominant; the ambiguity arises from the transcript itself rather than a clear model error.

& 

``The Daily Telegraph, one of the best in the MSM, at least had the grace to feature the brave women of Essex on its front page today. [...] \textbf{All Yvette Cooper and the Home Office are doing, stop polluting our country. \textcolor{red}{We don't just want illegals moved. We want them out.}} And what happened in Epping last night was actually quite wonderful. As the locals again defied the over-the-top police force and brought only peace, even though authorities tried to lock them in cages. [...] What will stop you? What will stop you? Nothing. \textbf{Nothing will stop us until, until, because it's our children. But when you say, oh yeah, your kids don't matter then, of course they matter.} \textbf{\textcolor{red}{Until they're safe, until the people that are unvetted are taken away.}} Once they're vetted, come back...... \textbf{They did too much to support Antifa on the first night and they \textcolor{red}{didn't protect us from them at all}. I'm really hoping this is to \textcolor{red}{protect us} from what they anticipated was coming from the left.} I'm really hoping that's what it's about. I can't actually imagine there are this many police to police a load of mums and dads walking to their local office. [...]'' 

& 
Political rivals act against the interests of the people

& 
\textbf{G:} Our borders are strained and/or insufficiently controlled \newline
\textbf{L:}  Immigration is a threat to individual safety
\\

\hline

\textbf{Failure to Detect Counter-Narratives (10\%).} The model fails to recognise that a narrative is being rejected, or corrected, and instead classifies the transcript as expressing that narrative.

&
`` \textcolor{red}{Since the government's policy of sending asylum seekers to Rwanda came into effect, I've noticed a lot of discussion about the supposed link between illegal migration and Britain's social housing shortage. But that link isn't quite as straightforward as some people would like to make it out to be. It's true that in England there are currently over 1 million households waiting for social housing, but so far in 2024 around 8,200 people have arrived illegally in Britain via small boats crossing the English Channel.} \textbf{They are not responsible for that enormous social housing waiting list.} And in fact, if you look at the government's own data, 90\% of all social housing lead tenants, that's the person whose name will be on the lease, are British. \textbf{So it's simply not true that all of our social housing is going to illegal migrants}  [...]''

&
None

&
\textbf{G:} Immigrants are a strain on the housing market \newline
\textbf{L:} Immigrants are a strain on the housing market
\\
\hline

\textbf{Weak Narrative Signal (5\%).} The transcript does not clearly express a strong narrative (often labelled as None), but the model predicts a narrative label based on weak topical cues or repeated keywords.

&
``[...] \textbf{They're going to start using TikTok influencers from Albania, Iraq, and Egypt and Iran. [...] There's been a million pounds set aside for this advertising campaign to get these influencers to encourage people to not try to cross over into the UK.} [...] \textbf{They're trying to tell people, look, there's nothing for you there... Don't go across the channel. It's very dangerous. You might drown. Whatever they're saying. And even when you get here, do you really want to join a machete drug gang?} [...]  \textcolor{red}{Because, of course, when people get here, they don't want to turn around and be like, oh, I made a terrible mistake. So they post all the pictures of themselves in the hotel rooms. Like we all do on Facebook. With the £10 notes. How great our life is going.}'' 

&

None

&

\textbf{G:} Immigrants are victims of false hope \newline
\textbf{L:} Immigrants are victims of false hope
\\
\hline
\end{tabular}

\caption{Error types and representative examples. Bold denotes our interpretation of the \textbf{gold-label rationale}, and red denotes our interpretation of \textcolor{red}{the model’s predicted-label rationale}. Both bold and red indicate the \textbf{\textcolor{red}{rationale for the multi-narrative overlap}}. Percentages are the frequency of each error type for both Llama (L) and Gemma (G). }
\label{tab:errors_examplesv2}

\end{table*}

\section{Error Analysis}

To better understand model limitations beyond overall performance, we conducted a qualitative error analysis on the \textbf{MigrationNarrate} development set, focusing on examples where both the top-performing open-source models, i.e., fine-tuned Llama and Gemma, misclassified the narrative label. In total, we manually inspected \textbf{60 video transcripts}, their gold label, and the predictions from both models. In examining the sources of errors, we grouped them into 5 recurring error types presented in Table~\ref{tab:errors_examplesv2}. These error types highlight both model limitations and the inherent complexities present in the labelled data.


Our analysis reveals systematic patterns underlying these errors. For instance, in \textit{Surface–Narrative Misalignment} cases, models tend to prioritise salient lexical cues over the dominant narrative framing. In Table~\ref{tab:errors_examplesv2}, row 1, the models predicted \textit{“There are too many immigrants
coming”} or \textit{“Our borders are strained and/or insufficiently controlled”}  based on explicit cues such as \texttt{“a record 973 illegal migrants entered Britain on 17 boats.” and “That is the biggest number
this year”}. However, the dominant narrative is \textit{“Political rivals’ policy failed/ is
failing/ will fail”}, as the video primarily emphasises that the Labour's plan to stop illegal immigrants is failing.
\textit{Fine-Grained Label Confusion} frequently arises in cases involving subtle differences in framing within the same narrative domain. For example, \textit{Receiving immigrants is too expensive/ immigration burdens
tax payers} and \textit{Natives loose to immigrants/ Immigrants receive better benefits than nationals} are highly related and often overlap semantically. As shown in Table~\ref{tab:errors_examplesv2}, row 2, although the primary focus is the financial burden on British taxpayers, both models incorrectly predicted the latter narrative because the video also mentions the free benefits provided to the migrants.
\textit{Multi-Narrative Overlap} reflects inherent ambiguity in the data, where multiple interpretations are plausible (Table~\ref{tab:errors_examplesv2}, row 3). Similarly, failures to detect counter-narratives indicate difficulty in capturing stance and negation, particularly in argumentative or conversational contexts (Table~\ref{tab:errors_examplesv2}, row 4). Finally, weak narrative signal cases reveal a tendency to over-predict narratives even in the absence of strong evidence. For instance, in Table~\ref{tab:errors_examplesv2}, row 5, although the gold label is \textit{None}, both models predicted \textit{Immigrants are victims of false
hope} likely to references to misleading portrayals of life in the UK on social media and unrealistic expectations. 

Overall, our findings reveal that improving narrative detection requires addressing reliance on surface cues and difficulty capturing implicit framing. \textit{Fine-Grained Label Confusion} and \textit{Multi-Narrative Overlap} highlight ambiguity, suggesting the need for clearer label boundaries in narrative taxonomies, as well as support for multi-label annotation.

\section{Conclusion}
In this paper, we presented \textbf{MigrationNarrate}, the first dataset for detecting migration narratives in videos transcripts. The dataset fills a critical research gap and provides a foundation for building tools for migration narrative detection. Additionally, we provided benchmark results by experimenting with a combination of pre-trained encoder models and both open- and closed-source LLMs. The results showed that fine-tuning open-source LLMs can outperform closed-source LLMs in zero-shot and few-shot settings at the super-narrative level; however, closed-source LLMs can still surpass them in fine-grained narrative detection. Finally, we conducted a thorough error analysis offering insights for future research directions.

\section*{Limitations}
\paragraph{Manual YouTube Channels and Search phrases selection} our data collection process relies on manually curated YouTube channels, playlists, and search phrases constructed based on the predefined super-narratives in the adopted JRC taxonomy. While this enables targeted retrieval of relevant content, it introduces potential selection bias and limits the representativeness of the dataset. In particular, super-narrative-driven query design may bias the data toward expected narrative frames, while channel- and playlist-level curation may further amplify certain viewpoints and filter out others.
\paragraph{\textbf{Annotators agreement}}
The dataset was annotated by only two annotators, with disagreements adjudicated by an expert on the task. Inter-annotator agreement is relatively low, reflecting the difficulty of the task and matching similar studies for narratives detection. In particular, the large number of narratives, the presence of overlapping narratives in some videos, and the requirement for annotators to assign a single dominant narrative increase task difficulty, leading to ambiguity and potential label noise.

\paragraph{\textbf{Absence of multimodal analysis}} Our approach is restricted to textual transcripts and does not account for video, audio, or other multimodal signals. This is a notable limitation, as narratives on YouTube are often conveyed through the interaction of visual, auditory, and textual elements. Consequently, our analysis may capture only a partial view of the underlying narratives. Future work should incorporate multimodal features, including visual content, audio cues, and user interactions, to provide a more comprehensive understanding.

\section*{Ethics Statement}
\paragraph{Ethics Review Board} Our research received the approval of a panel as required by our university Research Ethics Committee. Details are omitted in this version to preserve anonymity.
\paragraph{Annotators Consent and Compensation} We follow our university guidelines and research ethics guidelines and inform annotators of potential risks associated with the task, including exposure to sensitive or potentially harmful content. Annotators are given time to review the task details and consent form before participation and were informed that they are free to opt out at any point. The consent form provided to the annotators is presented in Appendix~\ref{consent}.
Additionally, participants received an hourly rate equivalent to the work of Graduate Teaching Assistants, in line with the University guidelines.
\paragraph{YouTube Terms and CopyRight}
The dataset is constructed from publicly available YouTube content in accordance with the platform’s Terms of Service.\footnote{\url{https://www.youtube.com/static?template=terms}} We do not redistribute audio, video, or transcripts. Instead, we provide only video identifiers and annotations, allowing the original content to be accessed through YouTube. We release the code to download the videos and extract the video transcripts with our data. The dataset is intended for research use only and will be released under a CC-BY-NC-SA 4.0 license.  
\paragraph{Privacy and Reputational Considerations}
The dataset is derived from publicly available YouTube content and does not include private or sensitive personal information. We restrict our study to textual transcripts and do not release audio or video data to reduce potential exposure of sensitive content. We also acknowledge a potential risk of reputational harm, as the dataset involves narratives that may be sensitive. However, our annotations focus on identifying narrative categories rather than evaluating the intent or credibility of individual speakers. Therefore, the dataset should not be interpreted as making claims about specific individuals.
\bibliographystyle{acl_natbib}
\bibliography{anthology,custom}

@misc{resourcera2026youtube,
  author       = {{Resourcera}},
  title        = {YouTube Statistics 2026: Users, Usage, and Key Metrics},
  year         = {2026},
  howpublished = {\url{https://resourcera.com/data/social/youtube-statistics/}},
  note         = {Accessed: April 15, 2026}
}

@inproceedings{salman-etal-2023-detecting,
    title = "Detecting Propaganda Techniques in Code-Switched Social Media Text",
    author = "Salman, Muhammad Umar  and
      Hanif, Asif  and
      Shehata, Shady  and
      Nakov, Preslav",
    editor = "Bouamor, Houda  and
      Pino, Juan  and
      Bali, Kalika",
    booktitle = "Proceedings of the 2023 Conference on Empirical Methods in Natural Language Processing",
    month = dec,
    year = "2023",
    address = "Singapore",
    publisher = "Association for Computational Linguistics",
    url = "https://aclanthology.org/2023.emnlp-main.1044/",
    doi = "10.18653/v1/2023.emnlp-main.1044",
    pages = "16794--16812"
}

@inproceedings{maarouf-etal-2024-hqp,
    title = "{HQP}: A Human-Annotated Dataset for Detecting Online Propaganda",
    author = {Maarouf, Abdurahman  and
      B{\"a}r, Dominik  and
      Geissler, Dominique  and
      Feuerriegel, Stefan},
    editor = "Ku, Lun-Wei  and
      Martins, Andre  and
      Srikumar, Vivek",
    booktitle = "Findings of the Association for Computational Linguistics: ACL 2024",
    month = aug,
    year = "2024",
    address = "Bangkok, Thailand",
    publisher = "Association for Computational Linguistics",
    url = "https://aclanthology.org/2024.findings-acl.363/",
    doi = "10.18653/v1/2024.findings-acl.363",
    pages = "6064--6089"
}

@inproceedings{wilby-etal-2023-gate,
    title = "{GATE} Teamware 2: An open-source tool for collaborative document classification annotation",
    author = "Wilby, David  and
      Karmakharm, Twin  and
      Roberts, Ian  and
      Song, Xingyi  and
      Bontcheva, Kalina",
    editor = "Croce, Danilo  and
      Soldaini, Luca",
    booktitle = "Proceedings of the 17th Conference of the European Chapter of the Association for Computational Linguistics: System Demonstrations",
    month = may,
    year = "2023",
    address = "Dubrovnik, Croatia",
    publisher = "Association for Computational Linguistics",
    url = "https://aclanthology.org/2023.eacl-demo.17/",
    doi = "10.18653/v1/2023.eacl-demo.17",
    pages = "145--151"
}

@inproceedings{piskorski-etal-2025-semeval,
    title = "{S}em{E}val 2025 Task 10: Multilingual Characterization and Extraction of Narratives from Online News",
    author = "Piskorski, Jakub  and
      Mahmoud, Tarek  and
      Nikolaidis, Nikolaos  and
      Campos, Ricardo  and
      Mario Jorge, Alipio  and
      Dimitrov, Dimitar  and
      Silvano, Purifica{\c{c}}{\~a}o  and
      Yangarber, Roman  and
      Sharma, Shivam  and
      Chakraborty, Tanmoy  and
      Guimaraes, Nuno  and
      Sartori, Elisa  and
      Stefanovitch, Nicolas  and
      Xie, Zhuohan  and
      Nakov, Preslav  and
      Da San Martino, Giovanni",
    editor = "Rosenthal, Sara  and
      Ros{\'a}, Aiala  and
      Ghosh, Debanjan  and
      Zampieri, Marcos",
    booktitle = "Proceedings of the 19th International Workshop on Semantic Evaluation (SemEval-2025)",
    month = jul,
    year = "2025",
    address = "Vienna, Austria",
    publisher = "Association for Computational Linguistics",
    url = "https://aclanthology.org/2025.semeval-1.331/",
    pages = "2610--2643",
    ISBN = "979-8-89176-273-2"
}

@inproceedings{ai2024tweetintent,
  title={TweetIntent@Crisis: A Dataset Revealing Narratives of Both Sides in the Russia-Ukraine Crisis},
  author={Ai, Lin and Gupta, Sameer and Oak, Shreya and Hui, Zheng and Liu, Zizhou and Hirschberg, Julia},
  booktitle={Proceedings of the International AAAI Conference on Web and Social Media},
  volume={18},
  pages={1872--1887},
  year={2024}
}

@article{wang2022text,
  title={Text embeddings by weakly-supervised contrastive pre-training},
  author={Wang, Liang and Yang, Nan and Huang, Xiaolong and Jiao, Binxing and Yang, Linjun and Jiang, Daxin and Majumder, Rangan and Wei, Furu},
  journal={arXiv preprint arXiv:2212.03533},
  year={2022}
}

@inproceedings{nikolaidis2025polynarrative,
  title={PolyNarrative: A Multilingual, Multilabel, Multi-domain Dataset for Narrative Extraction from News Articles},
  author={Nikolaidis, Nikolaos and Stefanovitch, Nicolas and Silvano, Purifica{\c{c}}{\~a}o and Dimitrov, Dimitar Iliyanov and Yangarber, Roman and Guimar{\~a}es, Nuno and Sartori, Elisa and Androutsopoulos, Ion and Nakov, Preslav and Da San Martino, Giovanni and others},
  booktitle={Proceedings of the 63rd Annual Meeting of the Association for Computational Linguistics (Volume 1: Long Papers)},
  pages={31323--31345},
  year={2025}
}

@inproceedings{haouari2025ukelectionnarratives,
  title={{UKElectionNarratives}: {A D}ataset of {M}isleading {N}arratives {S}urrounding {R}ecent {UK} {G}eneral {E}lections},
  author={Haouari, Fatima and Scarton, Carolina and Faggiani, Nicolo and Nikolaidis, Nikolaos and Kotseva, Bonka and Farha, Ibrahim Abu and Linge, Jens and Bontcheva, Kalina},
  booktitle={Proceedings of the International AAAI Conference on Web and Social Media},
  volume={19},
  pages={2477--2495},
  year={2025}
}

@article{liu2019roberta,
  title={Roberta: A robustly optimized bert pretraining approach},
  author={Liu, Yinhan and Ott, Myle and Goyal, Naman and Du, Jingfei and Joshi, Mandar and Chen, Danqi and Levy, Omer and Lewis, Mike and Zettlemoyer, Luke and Stoyanov, Veselin},
  journal={arXiv preprint arXiv:1907.11692},
  year={2019}
}

@misc{grattafiori2024llama3herdmodels,
      title={The Llama 3 Herd of Models}, 
      author={Aaron Grattafiori and Abhimanyu Dubey and Abhinav Jauhri and Abhinav Pandey and Abhishek Kadian and Ahmad Al-Dahle and Aiesha Letman and Akhil Mathur and Alan Schelten and Alex Vaughan and 1 others},
      year={2024},
      eprint={2407.21783},
      archivePrefix={arXiv},
      primaryClass={cs.AI},
      url={https://arxiv.org/abs/2407.21783}, 
}

@misc{gemmateam2025gemma3technicalreport,
      title={Gemma 3 Technical Report}, 
      author={Aishwarya Kamath and Johan Ferret and Shreya Pathak and Nino Vieillard and Ramona Merhej and Sarah Perrin and Tatiana Matejovicova and Alexandre Ramé and Morgane Rivière and Louis Rouillard and Thomas Mesnard and Geoffrey Cideron and 1 others},
      year={2025},
      eprint={2503.19786},
      archivePrefix={arXiv},
      primaryClass={cs.CL},
      url={https://arxiv.org/abs/2503.19786}, 
}

@misc{qwen2025qwen25technicalreport,
      title={Qwen2.5 Technical Report}, 
      author={An Yang and Baosong Yang and Beichen Zhang and Binyuan Hui and Bo Zheng and Bowen Yu and Chengyuan Li and Dayiheng Liu and 1 others},
      year={2025},
      eprint={2412.15115},
      archivePrefix={arXiv},
      primaryClass={cs.CL},
      url={https://arxiv.org/abs/2412.15115}, 
}

@inproceedings{singh2025gatenlp,
  title={GateNLP at SemEval-2025 task 10: Hierarchical three-step prompting for multilingual narrative classification},
  author={Singh, Iknoor and Scarton, Carolina and Bontcheva, Kalina},
  booktitle={Proceedings of the 19th International Workshop on Semantic Evaluation (SemEval-2025)},
  pages={148--154},
  year={2025}
}

@article{dettmers2023qlora,
  title={Qlora: Efficient finetuning of quantized llms},
  author={Dettmers, Tim and Pagnoni, Artidoro and Holtzman, Ari and Zettlemoyer, Luke},
  journal={Advances in neural information processing systems},
  volume={36},
  pages={10088--10115},
  year={2023}
}

@article{singh2025openai,
  title={Openai gpt-5 system card},
  author={Singh, Aaditya and Fry, Adam and Perelman, Adam and Tart, Adam and Ganesh, Adi and El-Kishky, Ahmed and McLaughlin, Aidan and Low, Aiden and Ostrow, AJ and Ananthram, Akhila and others},
  journal={arXiv preprint arXiv:2601.03267},
  year={2025}
}

@article{ghafouri2025framing,
  title={Framing migration: A computational analysis of uk parliamentary discourse},
  author={Ghafouri, Vahid and McNeil, Robert and Yankov, Teodor and Sumption, Madeleine and Rocher, Luc and Hale, Scott A and Mahdi, Adam},
  journal={arXiv preprint arXiv:2509.14197},
  year={2025}
}

@inproceedings{dingemanse-liesenfeld-2022-text,
    title = "From text to talk: {H}arnessing conversational corpora for humane and diversity-aware language technology",
    author = "Dingemanse, Mark  and
      Liesenfeld, Andreas",
    editor = "Muresan, Smaranda  and
      Nakov, Preslav  and
      Villavicencio, Aline",
    booktitle = "Proceedings of the 60th Annual Meeting of the Association for Computational Linguistics (Volume 1: Long Papers)",
    month = may,
    year = "2022",
    address = "Dublin, Ireland",
    publisher = "Association for Computational Linguistics",
    url = "https://aclanthology.org/2022.acl-long.385/",
    doi = "10.18653/v1/2022.acl-long.385",
    pages = "5614--5633"
}

@article{sahin2020migration,
  title={Migration narratives in policy and politics},
  author={Sahin-Mencutek, Zeynep},
  journal={Retrieved January},
  volume={22},
  pages={2021},
  year={2020}
}

@misc{garces2025making,
  title={The making of migration narratives: understanding processes and gauging impacts},
  author={Garc{\'e}s-Mascare{\~n}as, Blanca and Pastore, Ferruccio},
  journal={Journal of Ethnic and Migration Studies},
  volume={51},
  number={16},
  pages={4119--4137},
  year={2025},
  publisher={Taylor \& Francis}
}

@article{smellie2025influence,
  title={The influence of media narratives on political debate: narratives on the 2015 migrant ‘crisis’ in five European countries},
  author={Smellie, Saskia and Boswell, Christina},
  journal={Journal of Ethnic and Migration Studies},
  volume={51},
  number={16},
  pages={4181--4200},
  year={2025},
  publisher={Taylor \& Francis}
}

@article{maneri2023comparative,
  title={A comparative analysis of migration narratives in traditional and social media},
  author={Maneri, Marcello and others},
  journal={Bridges},
  pages={1--84},
  year={2023},
  publisher={Zenodo}
}

@article{dennison2021narratives,
  title={Narratives: a review of concepts, determinants, effects, and uses in migration research},
  author={Dennison, James},
  journal={Comparative Migration Studies},
  volume={9},
  number={1},
  pages={50},
  year={2021},
  publisher={Springer}
}

@misc{openai2026gpt54,
  author       = {{OpenAI}},
  title        = {Introducing GPT-5.4},
  year         = {2026},
  howpublished = {\url{https://openai.com/index/introducing-gpt-5-4/}},
  note         = {Accessed: 2026-04-26}
}

@inproceedings{chen2024open,
  title={Open-set semi-supervised text classification via adversarial disagreement maximization},
  author={Chen, Junfan and Zhang, Richong and Chen, Junchi and Hu, Chunming},
  booktitle={Proceedings of the 62nd Annual Meeting of the Association for Computational Linguistics (Volume 1: Long Papers)},
  pages={2170--2180},
  year={2024}
}

@inproceedings{yang2025calibrating,
  title={Calibrating Pseudo-Labeling with Class Distribution for Semi-supervised Text Classification},
  author={Yang, Weiyi and Zhang, Richong and Chen, Junfan and Sheng, Jiawei},
  booktitle={Proceedings of the 2025 Conference on Empirical Methods in Natural Language Processing},
  pages={13026--13039},
  year={2025}
}

@article{hurst2024gpt,
  title={{GPT-4o} system card},
  author={Hurst, Aaron and Lerer, Adam and Goucher, Adam P and Perelman, Adam and Ramesh, Aditya and Clark, Aidan and Ostrow, AJ and Welihinda, Akila and Hayes, Alan and Radford, Alec and others},
  journal={arXiv preprint arXiv:2410.21276},
  year={2024}
}

@inproceedings{heinrich-etal-2024-automatic,
    title = "Automatic Identification of {COVID}-19-Related Conspiracy Narratives in {G}erman Telegram Channels and Chats",
    author = {Heinrich, Philipp  and
      Blombach, Andreas  and
      Doan Dang, Bao Minh  and
      Zilio, Leonardo  and
      Havenstein, Linda  and
      Dykes, Nathan  and
      Evert, Stephanie  and
      Sch{\"a}fer, Fabian},
    editor = "Calzolari, Nicoletta  and
      Kan, Min-Yen  and
      Hoste, Veronique  and
      Lenci, Alessandro  and
      Sakti, Sakriani  and
      Xue, Nianwen",
    booktitle = "Proceedings of the 2024 Joint International Conference on Computational Linguistics, Language Resources and Evaluation (LREC-COLING 2024)",
    month = may,
    year = "2024",
}

@inproceedings{piskorski2022exploring,
  title={Exploring Data Augmentation for Classification of Climate Change Denial: Preliminary Study.},
  author={Piskorski, Jakub and Nikolaidis, Nikolaos and Stefanovitch, Nicolas and Kotseva, Bonka and Vianini, Irene and Kharazi, Sopho and Linge, Jens P and others},
  booktitle={Text2Story@ ECIR},
  pages={97--109},
  year={2022}
}

@inproceedings{sechidis2011stratification,
  title={On the stratification of multi-label data},
  author={Sechidis, Konstantinos and Tsoumakas, Grigorios and Vlahavas, Ioannis},
  booktitle={ ECML PKDD 2011},
  year={2011}
}

@article{fraile2024automatic,
  title={Automatic Identification of Narratives: Evaluation framework, annotation methodology and dataset creation},
  author={Fraile-Hern{\'a}ndez, Jes{\'u}s M and Pe{\~n}as, Anselmo and Moral, Pablo},
  journal={IEEE Access},
  year={2024},
  publisher={IEEE}
}

@article{kotseva2023trend,
  title={Trend analysis of COVID-19 mis/disinformation narratives--A 3-year study},
  author={Kotseva, Bonka and Vianini, Irene and Nikolaidis, Nikolaos and Faggiani, Nicol{\`o} and Potapova, Kristina and Gasparro, Caroline and Steiner, Yaniv and Scornavacche, Jessica and Jacquet, Guillaume and Dragu, Vlad and others},
  journal={Plos one},
  volume={18},
  number={11},
  pages={e0291423},
  year={2023},
  publisher={Public Library of Science San Francisco, CA USA}
}

@article{coan2021computer,
  title={Computer-assisted classification of contrarian claims about climate change},
  author={Coan, Travis G and Boussalis, Constantine and Cook, John and Nanko, Mirjam O},
  journal={Scientific reports},
  volume={11},
  number={1},
  pages={22320},
  year={2021},
  publisher={Nature Publishing Group UK London}
}

@inproceedings{rowlands-etal-2024-predicting,
    title = "Predicting Narratives of Climate Obstruction in Social Media Advertising",
    author = "Rowlands, Harri  and
      Morio, Gaku  and
      Tanner, Dylan  and
      Manning, Christopher",
    editor = "Ku, Lun-Wei  and
      Martins, Andre  and
      Srikumar, Vivek",
    booktitle = "ACL 2024",
    month = aug,
    year = "2024",
}

@techreport{JRC142039,
	number = {KJ-01-25-336-EN-N (online),KJ-01-25-336-EN-C (print)},
	address = {Luxembourg (Luxembourg)},
	issn = {1831-9424 (online),1018-5593 (print)},
	year = {2025},
	author = {Fiona Seiger and Nina Kajander and Alberto-Horst Neidhardt and Mario Scharfbillig and Lenka Dražanová and Christoph Deuster and Michal Krawczyk and Andrea Blasco and Rossella Icardi and Marina Tzvetkova and Lieke Bakker and Karelis Olivo Rumpf},
	isbn = {978-92-68-28628-9 (online),978-92-68-28629-6 (print)},
	publisher = {Publications Office of the European Union},

	title = {Navigating migration narratives}}

@inproceedings{radford2023robust,
  title={Robust speech recognition via large-scale weak supervision},
  author={Radford, Alec and Kim, Jong Wook and Xu, Tao and Brockman, Greg and McLeavey, Christine and Sutskever, Ilya},
  booktitle={Proceedings of the 40th International Conference on Machine Learning},
  pages={28492--28518},
  year={2023}
}

\appendix

\section{Narrative Definitions}
\label{sec:narratives_def}

In this section, we present the narrative definitions used in our annotation guidelines and in our experiments.
For readability, each narrative definition is presented using the format \textbf{super-narrative/\textcolor{blue}{narrative}: narrative definition}, allowing readers to easily identify the corresponding super-narrative. Additionally we present our prompt to GPT-5.1 to get the initial versions of our narrative definitions before manually reviewing them.

\begin{tcolorbox}[
  enhanced,
  colback=white,
  colframe=black,
  coltitle=black,
  colbacktitle=gray!50,
  title=\tiny\textbf{Narrative Definitions Prompt},
  fontupper=\tiny,
  boxrule=1pt,
  arc=5pt,
  left=8pt,
  right=8pt,
  top=8pt,
  bottom=7pt
]

You are a narrative-taxonomy assistant. Your task is to write clear, concise, and non-overlapping definitions for narratives that belong to a broader immigration super-narrative.

\medskip
\textbf{Inputs:}
\begin{itemize}\setlength{\itemsep}{2pt}
  \item Super-narrative: \texttt{<super-narrative>}
  \item Super-narrative definition: \texttt{<super-narrative definition>}
  \item List of narratives: \texttt{<comma-separated list of narratives>}
\end{itemize}

\medskip
\textbf{Instructions:}
\begin{enumerate}\setlength{\itemsep}{2pt}
  \item For each narrative, write 1--3 sentences defining it.
  \item Definitions must fit logically under the super-narrative but be conceptually distinct.
  \item Describe what each narrative claims without expressing personal, ethical, or political judgments.
\end{enumerate}

\medskip
\textbf{Output format:}
\begin{verbatim}
{
  "super_narrative": "<super-narrative>",
  "narrative_definitions": {
    "<narrative_1>": "<definition>",
    "<narrative_2>": "<definition>"
    .....
  }
}
\end{verbatim}

\end{tcolorbox}

\begin{enumerate}
    \item \textbf{Immigrants’ identity and/or culture is problematic/\textcolor{blue}{Immigration is a threat to the European way of life / identity}:} This narrative claims that immigration introduces cultural changes that weaken or replace long-standing European customs, values, and collective identity. It asserts that demographic and cultural shifts driven by immigration risk undermining a cohesive and historically rooted European social fabric.
    \item \textbf{Immigrants’ identity and/or culture is problematic/\textcolor{blue}{Arabs and/or Muslims are a cultural/social threat}:} This narrative portrays Arabs and/or Muslims as holding cultural or religious practices that are fundamentally incompatible with European norms and values. It suggests that their presence disrupts social cohesion and challenges culturally accepted ways of life associated with European societies.
    \item \textbf{Immigrants’ identity and/or culture is problematic/\textcolor{blue}{Certain immigrants are unwilling/incapable to integrate}:} This narrative argues that some immigrants do not adopt the host society’s language, values, and cultural expectations, or are inherently unable to do so. It interprets continued cultural distinctions as evidence that meaningful integration into European societies is unlikely to occur for these groups.
    \item \textbf{Immigration is a threat/\textcolor{blue}{Immigration is a threat to individual safety}:} This narrative claims that the presence of immigrants increases risks to the physical safety and daily security of individuals. It suggests that ordinary citizens become more vulnerable to harm due to immigration.
    \item \textbf{Immigration is a threat/\textcolor{blue}{Immigration is a threat to national security}:} This narrative asserts that immigration undermines the safety and stability of the state by allowing potentially dangerous individuals or groups to enter the country. It frames immigration as a vulnerability that can be exploited to compromise national defence and security systems.
     \item \textbf{Immigration is a threat/\textcolor{blue}{Immigrants are prone to committing crimes (violent, non-violent, or organised)}:} This narrative portrays immigrants as more likely than non-immigrants to engage in criminal activity, whether petty, violent, or organised crime. It claims that immigration increases crime rates and strains law enforcement systems.
    \item \textbf{Immigration is a threat/\textcolor{blue}{Immigrants are prone to committing crimes of a sexual nature}:} This narrative alleges that immigrant men, in particular, are more likely to commit sexual harassment, assault, or rape. It characterises immigration as heightening risks of sexually motivated harm to women and children.
    \item \textbf{Immigration is a threat/\textcolor{blue}{Many immigrants are terrorists}:} This narrative claims that immigration allows terrorists or radicalised individuals to enter the country and conduct attacks. It suggests that immigration is a direct pathway for extremist violence and instability.
   \item \textbf{Immigration is a threat/\textcolor{blue}{Immigrants spread diseases}:} This narrative asserts that immigrants are carriers of infectious diseases that pose health risks to the native population. It frames immigration as a public health threat capable of triggering outbreaks or overwhelming healthcare systems.
   \item \textbf{Immigration is out of control/\textcolor{blue}{There are too many immigrants coming}:} This narrative claims that the volume of arriving immigrants exceeds what the country can manage socially, economically, or culturally. It suggests that the pace and scale of immigration are unsustainable and create a perception of overwhelming influx.
    \item \textbf{Immigration is out of control/\textcolor{blue}{Immigrants do not contribute to the economy}:} This narrative claims that immigrants add little or no value to the host economy. It suggests that they rely on public resources without generating sufficient tax revenue or economic growth in return.
    \item \textbf{Immigration burdens the economy \& welfare narrative/\textcolor{blue}{Our borders are strained and/or insufficiently controlled}:} This narrative asserts that border management systems are weak, unable to regulate who enters the country, or incapable of preventing irregular migration. It frames borders as overwhelmed or poorly enforced, implying that immigration is occurring without adequate oversight or restriction.
     \item \textbf{Immigration burdens the economy \& welfare narrative/\textcolor{blue}{Immigrants take our jobs}:} This narrative asserts that immigrants compete directly with native citizens for employment opportunities. It suggests that immigration reduces job availability or lowers wages for the native workforce.
    \item \textbf{Immigration burdens the economy \& welfare narrative/\textcolor{blue}{Many asylum seekers are actually economic migrants}:} This narrative claims that a significant number of asylum seekers are motivated primarily by financial gain rather than persecution. It portrays asylum claims as a strategic route to access economic benefits in the host country.
    \item \textbf{Immigration burdens the economy \& welfare narrative/\textcolor{blue}{Natives loose to immigrants / Immigrants receive better benefits than nationals}:} This narrative alleges that immigrants receive more favourable access to welfare, housing, employment, or social services than native-born citizens. It claims that natives are disadvantaged or deprioritised as a result.
    \item \textbf{Immigration burdens the economy \& welfare narrative/\textcolor{blue}{Immigrants abuse the welfare system}:} This narrative alleges that immigrants receive more favourable access to welfare, housing, employment, or social services than native-born citizens. It claims that natives are disadvantaged or deprioritised as a result. 
    \item \textbf{Immigration burdens the economy \& welfare narrative/\textcolor{blue}{Immigrants are a strain on our health-care system}:} This narrative claims that immigration increases pressure on healthcare services, making it harder for native citizens to access timely and affordable medical care. It frames immigrants as consuming resources beyond the system’s capacity.
    \item \textbf{Immigration burdens the economy \& welfare narrative/\textcolor{blue}{Immigrants are a strain on the housing market}:} This narrative asserts that immigration contributes to housing shortages, rising rents, or limited access to affordable accommodation for native residents. It suggests that increased demand caused by immigration intensifies competition for housing.
    \item \textbf{Immigration burdens the economy \& welfare narrative/\textcolor{blue}{Natives first}:} This narrative claims that economic resources, welfare services, and public support should prioritise native-born citizens over immigrants. It promotes the idea that natives are entitled to preferential treatment due to their status or historical connection to the nation.
    \item \textbf{Immigration burdens the economy \& welfare narrative/\textcolor{blue}{Immigration reduces the attractiveness of location and lowers overall life quality}:} This narrative suggests that immigration negatively affects local communities by reducing standards of living, safety, or neighbourhood desirability. It implies that immigration leads to declining quality of public services, infrastructure, or community well-being.
    \item \textbf{Immigration burdens the economy \& welfare narrative/\textcolor{blue}{Receiving immigrants is too expensive/immigration burdens tax payers}:} This narrative asserts that the financial costs of immigration—such as welfare, public services, integration programmes, and housing—place an excessive burden on taxpayers. It frames immigration as economically unsustainable due to the high level of public spending it requires.
    \item \textbf{Immigrants are victims/\textcolor{blue}{Immigrants are victims of false hope}:} This narrative claims that immigrants are misled into believing that life in Europe will guarantee security, prosperity, or opportunity, although these expectations are unlikely to be fulfilled. It presents migrants as suffering because they acted on unrealistic promises or inaccurate information about their prospects abroad.
    \item \textbf{Immigrants are victims/\textcolor{blue}{Immigrants are victims of human traffickers}:} This narrative asserts that migrants fall prey to criminal networks that exploit their vulnerability for profit. It claims that traffickers manipulate, coerce, or deceive immigrants, exposing them to dangerous travel routes, abuse, or exploitation.

  \item \textbf{Immigrants are victims/\textcolor{blue}{EU/MS do not have the means to properly take care of immigrants, therefore it is better for them not to come here}:} This narrative argues that European institutions or member states lack the capacity to provide adequate support, services, or protection to immigrants. It presents the inability to ensure proper living conditions as a reason why migrants ultimately suffer by coming.

  \item \textbf{Immigrants are victims/\textcolor{blue}{Certain immigrants /asylum seekers are discriminated against}:} This narrative claims that specific groups of immigrants or asylum seekers are subjected to unequal treatment due to their nationality, ethnicity, religion, or other characteristics. It frames institutional or social practices as systematically disadvantaging these migrants.
  
  \item \textbf{Immigrants are victims/\textcolor{blue}{MS or EU are in breach of international law (e.g. inadequate reception facilities, refoulement at the borders, prolonged asylum procedures)}:} This narrative asserts that European institutions or member states violate international or humanitarian law in their treatment of migrants. It highlights failures such as unlawful pushbacks, substandard living conditions, or excessively long asylum procedures as evidence of wrongdoing.

\item \textbf{Immigrants are victims/\textcolor{blue}{Immigrants suffer from labour exploitation}:} This narrative claims that immigrants experience unfair or abusive working conditions, including low wages, excessive hours, or unsafe environments. It presents migrants as vulnerable to economic exploitation due to their precarious legal or social position in the host country.

\item \textbf{Immigration is part of a conspiracy/\textcolor{blue}{Blaming global elites}:} This narrative claims that powerful political, economic, or institutional actors deliberately promote or engineer migration flows for hidden strategic purposes. It suggests that immigration is not a natural social process but the result of a coordinated agenda imposed from above.

\item \textbf{Immigration is part of a conspiracy/\textcolor{blue}{Great replacement}:} This narrative asserts that there is a deliberate plan to replace native European populations with non-European immigrants. It frames demographic change as intentional and directed, with the aim of eroding cultural, ethnic, or national identity.

\item \textbf{Immigration is part of a conspiracy/\textcolor{blue}{Other conspiracy theory}:} This narrative applies conspiracy-based explanations to immigration beyond the ideas of global elite orchestration or population replacement. It claims that hidden actors or covert agendas are responsible for migration patterns, policies, or crises rather than social, economic, or geopolitical factors.

\item \textbf{Pragmatic approaches to immigration/\textcolor{blue}{Immigration helps tackle social \& economic issues (e.g. skills and labour shortages)}:} This narrative claims that immigration provides practical benefits by filling critical gaps in the labour market, addressing demographic challenges, or supporting essential public services. It suggests that immigration can strengthen the host society when aligned with specific economic or social needs.

\item \textbf{Geopolitical narratives: Countries are acting against the interest of other countries/\textcolor{blue}{Immigration is abused as a political tool}:} This narrative claims that governments or geopolitical actors intentionally manipulate migration flows to pressure, destabilise, or weaken other states. It portrays immigration policy and population movement as weapons used strategically in international competition rather than as neutral or humanitarian matters.

\item \textbf{Geopolitical narratives: Countries are acting against the interest of other countries/\textcolor{blue}{Our country is doing more than other countries to deal with immigration issues}:} This narrative asserts that the home country carries a disproportionate burden in managing migration while other states fail to share responsibility or take adequate action. It frames the imbalance as strategically disadvantageous and suggests that the country is disadvantaged in the geopolitical landscape as a result.

\item \textbf{We are competent /\textcolor{blue}{Promoting policy initiatives or ideas}:} This narrative claims that the political actor or party has concrete and effective proposals to address immigration challenges. It presents policy plans as practical solutions that will fix perceived problems and strengthen control over immigration.

\item \textbf{We are competent/\textcolor{blue}{Applauding political allies}:} This narrative asserts that other domestic or international political actors who share similar views on immigration are capable and trustworthy partners. It highlights these alliances as evidence of strong leadership and as reinforcement of the party’s credibility.

\item \textbf{We are competent/\textcolor{blue}{We represent the people}:} This narrative claims that the political actor or party defends the interests and concerns of ordinary citizens regarding immigration. It positions the party as the only legitimate voice of the population against elites or institutions that allegedly ignore public demands.

\item \textbf{We are competent/\textcolor{blue}{We stand for common sense}:} This narrative asserts that the party’s immigration views reflect basic logic, practicality, and everyday reasoning. It implies that alternative migration approaches are irrational or disconnected from the lived realities of ordinary people.

\item \textbf{We are competent/\textcolor{blue}{We will reinstate order}:} This narrative claims that the political actor or party is capable of restoring control and stability in relation to immigration. It presents the party as able to correct mismanagement, curb chaos, and impose clear rules to ensure societal security and predictability.

\item \textbf{It is Us vs. Them (the establishment)/\textcolor{blue}{Discrediting liberal values}:} This narrative claims that liberal ideals—such as multiculturalism, human rights, or open societies—are harmful to the nation and its people. It frames these values as imposed by the elite and opposed to the interests of ordinary citizens.

\item \textbf{It is Us vs. Them (the establishment)/\textcolor{blue}{EU is harmful for our nation}:} This narrative asserts that the European Union undermines national interests by imposing regulations, policies, or migration rules that weaken the country. It portrays EU membership as limiting national autonomy and harming the well-being of citizens.

\item \textbf{It is Us vs. Them (the establishment)/\textcolor{blue}{Let’s overthrow the establishment}:} This narrative calls for the removal of the existing political leadership or governing institutions that are depicted as betraying public interests. It frames radical political change as necessary to regain control over national decision-making, especially regarding migration.

\item \textbf{It is Us vs. Them (the establishment)/\textcolor{blue}{Our sovereignty is under threat}:} This narrative claims that national sovereignty is being eroded by international institutions, supranational actors, or global elites. It suggests that external forces dictate migration policies, preventing the nation from acting independently in the interests of its citizens.

\item \textbf{It is Us vs. Them (the establishment)/\textcolor{blue}{Discrediting political rivals (e.g. ridicule, accusation)}:} This narrative attacks political opponents by portraying them as incompetent, corrupt, foolish, or dishonest. It positions rivals as unfit to govern, especially in relation to migration, and emphasizes their alleged failures or ulterior motives.

\item \textbf{It is Us vs. Them (the establishment)/\textcolor{blue}{Political rivals’ policy failed/ is failing/ will fail}:} This narrative claims that policies supported by political opponents have already failed, are currently failing, or are destined to fail. It highlights perceived negative outcomes of their migration policies as evidence that rivals cannot manage the issue effectively.

\item \textbf{It is Us vs. Them (the establishment)/\textcolor{blue}{Political rivals act against the interests of
the people}:} This narrative asserts that political opponents represent elites, foreigners, or special interest groups rather than ordinary citizens. It claims that rivals intentionally ignore or undermine the needs and concerns of the population regarding migration.

\item \textbf{It is Us vs. Them (the establishment)/\textcolor{blue}{Vindication}:} This narrative claims that past warnings or predictions made by the populist party or leader about migration have proven correct. It presents unfolding events as confirmation that the party’s position has always been accurate, reinforcing trust in its leadership.

\item \textbf{It is Us vs. Them (the establishment)/\textcolor{blue}{Anti-Elitism}:} This narrative portrays elites—political, financial, academic, or media—as self-serving actors disconnected from ordinary people. It claims that the elite deliberately impose migration policies that harm the public while benefiting themselves or their networks.

\item \textbf{Narrative strategy to remain in the news/\textcolor{blue}{Reminder that immigration is an issue}:} This narrative claims that immigration must remain a central topic of public concern and debate. It repeatedly emphasizes that immigration is a pressing, unresolved, or urgent problem to ensure continuous media coverage and public attention.

\item \textbf{Narrative strategy to remain in the news/\textcolor{blue}{Self-promotion}:} This narrative asserts that the political actor or party positions itself as highly relevant, visible, and central in discussions on immigration. It highlights their role, actions, or statements as newsworthy to maintain prominence in the media landscape.

\item \textbf{There is an unfair bias against our political camp/\textcolor{blue}{We are being silenced}:} This narrative claims that the political group is deliberately prevented from expressing its views on immigration or participating fully in public debate. It suggests that opponents or institutions actively suppress their opinions, visibility, or influence.

\item \textbf{There is an unfair bias against our political camp/\textcolor{blue}{Mainstream media are biased}:} This narrative asserts that major media outlets consistently portray the political group negatively or downplay its arguments. It suggests that journalists and media organisations intentionally favour rival political actors and distort reporting on immigration.

\item \textbf{There is an unfair bias against our political camp/\textcolor{blue}{The judicial system favours political rivals and/or obstructs own political camp}:} This narrative claims that courts, prosecutors, or legal institutions apply the law unevenly, shielding political opponents while targeting or hindering one’s own political group. It describes legal decisions as part of a coordinated effort to disadvantage that group, especially on immigration matters.

\item \textbf{There is an unfair bias against our political camp/\textcolor{blue}{There are double-standards/ the system is corrupt}:} This narrative asserts that political, legal, or institutional systems operate with biased rules that benefit rival groups and penalise one's own political camp. It frames corruption or partiality as intentional and systemic, reinforcing the idea of an unfair playing field.
\end{enumerate}


\section{Videos Collection Source}\label{appendix:videos_collection}
In Table~\ref{tab:youtube_channels}, we present the YouTube channels and playlists from which we collected our data. 
\begin{table}[!htb]
\tiny
\centering
\begin{tabular}{ll}
\hline
\textbf{YouTube Channel} & \textbf{YouTube Playlist} \\
\hline
@mwukofficial& All videos\\
@ReformUKOfficial& All videos\\
@labourparty& All videos\\
@Conservatives& All videos\\
@ukhomeoffice& All videos\\
@NigelFarageOfficial& All videos\\
@KeirStarmer& All videos\\
@TommyRobinsonOnline& All videos\\
@JimmyTheGiant& All videos\\
@Bigtastytompall& All videos\\
@NewCultureForum& All videos\\
@drmatthewgoodwin& All videos\\
@GBNewsOnline & Farage \\
@GBNewsOnline & Patrick Christys Tonight \\
@GBNewsOnline & Britain's Newsroom \\
@GBNewsOnline & Dewbs \& Co \\
@BBCNews & UK \\
@BBCNews & Shorts \\
@SkyNews & Migration Crisis \\
@SkyNews & Politics 2025 \\
@SkyNews & UK News 2025\\
@SkyNews & Migration Crisis \\
@SkyNews & UK News \\
@SkyNews & UK Riots\\
@SkyNews & Sky News Daily Podcasts \\
@SkyNews & The UK Tonight with Sarah-Jane Mee \\
@SkyNews & Strikes\\
@channel4news & UK News \\
@channel4news& UK Politics\\
@ITVNews & News Shorts\\
@gmb& Shorts\\
@gmb& Headlines\\
@talktv & News Shorts\\
@talktv& Shorts\\
@guardiannews& Headlines\\
@DailyExpress& UK News\\
@DailyExpress& UK Politics\\

\hline
\end{tabular}

\caption{List of YouTube channels and corresponding playlists used for data collection.}
\label{tab:youtube_channels}
\end{table}

\section{Videos Search Phrases}\label{search_phrases}
We present in Table~\ref{tab:search_phrases} and Table~\ref{tab:narrative_search_phrases} the general and narrative\_oriented search phrases we used to search for videos respectively.
\begin{table}[h]
\centering
\small
\begin{tabular}{c}
\toprule
Immigration UK \\
UK Immigration \\
Immigration in the UK\\
Migrants UK\\
UK migrant crisis\\
UK migration crisis \\
Illegal immigration UK\\
Asylum UK \\
UK asylum seekers\\
Refugees UK\\
Refugee crisis UK\\
Channel crossings UK\\
Small boats UK\\
Stop the boats UK\\
Border control UK\\
\bottomrule
\end{tabular}
\caption{General search phrases.}
\label{tab:search_phrases}
\end{table}

\section{Semantic Filtering Queries}\label{appendix:queries}
We present below the list of queries we adopted for filtering the videos semantically:
\begin{enumerate}
    \item immigration in the UK
    \item migration to the UK
    \item migrants coming to or living in the UK
    \item refugees in the UK
    \item asylum seekers in the UK
    \item Channel crossings and small boats coming to the UK
    \item border control, deportations and visas in the UK
    \item illegal immigration and people smuggling to the UK
    \item government plans and deportation schemes for asylum seekers in the UK
    \item UK immigration policy, laws and the migration bill

\end{enumerate}

\section{Annotation Framework}\label{appendix:annotation_framework}
\begin{figure*}[!htbp]
\centering
\includegraphics[scale=0.75]{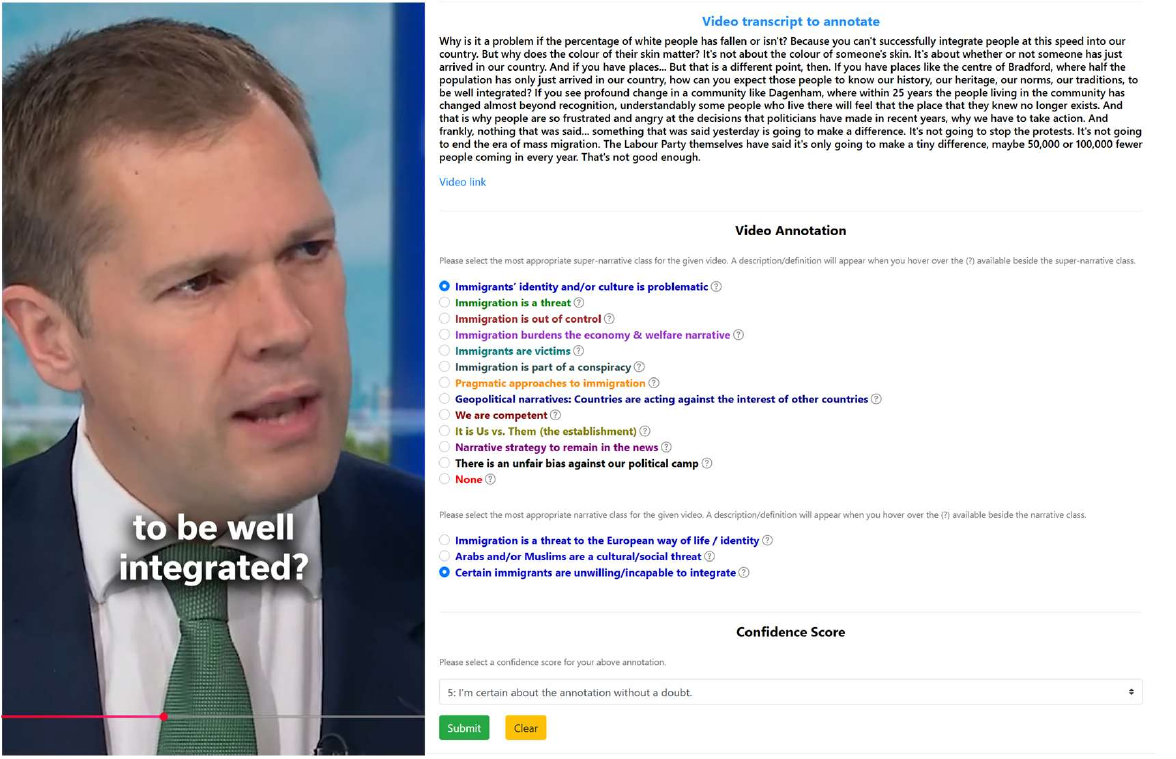}
\caption{Annotation Framework. The annotator is shown the video transcript along with a direct link to the YouTube video. First, they are presented with the list of super-narratives only. Once a super-narrative is selected, the corresponding list of narratives is displayed. Label definitions are displayed when the annotator hovers over them. }
\label{annotation_framework}
\end{figure*}

We present in Figure~\ref{annotation_framework}, the annotation tool as presented to the annotators.
\section{Benchmarking prompt}\label{appendix:benchmark_prompt}
\begin{tcolorbox}[
  enhanced,
  colback=white,
  colframe=black,
  coltitle=black,
  colbacktitle=gray!50,
  title=\tiny\textbf{LLMs Two-step Prompt (Step 1)},
  fontupper=\tiny,
  boxrule=1pt,
  arc=5pt,
  left=8pt,
  right=8pt,
  top=8pt,
  bottom=7pt
]

Classify the given video transcript into the most relevant migration super\_narrative.

\medskip
\textbf{Super\_narratives:}

super\_narrative 1: super\_narrative 1 definition

super\_narrative 2: super\_narrative 2 definition

....

....

\textbf{Examples:}

Video transcript: \{video transcript 1\}

Super\_narrative: \{super\_narrative 1\}

Video transcript: \{video transcript 2\}

Super\_narrative: \{super\_narrative 2\}

....

....

\medskip
Video transcript:
\medskip

Return the most relevant super\_narrative. If no super\_narrative applies, return ``None''. Return only the super\_narrative label without any additional explanation.

\end{tcolorbox}

\begin{tcolorbox}[
  enhanced,
  colback=white,
  colframe=black,
  coltitle=black,
  colbacktitle=gray!50,
  title=\tiny\textbf{LLMs Two-step Prompt (Step 2)},
  fontupper=\tiny,
  boxrule=1pt,
  arc=5pt,
  left=8pt,
  right=8pt,
  top=8pt,
  bottom=7pt
]

The given video transcript is related to migration, and its main super\_narrative is: \{super\_narrative\}.  Classify it into the most relevant narrative.

\medskip
\textbf{Narratives:}

narrative 1: narrative 1 definition

narrative 2: narrative 2 definition

....

....

\textbf{Examples:}

Video transcript: \{video transcript 1\}

Narrative: \{Narrative 1\}

Video transcript: \{video transcript 2\}

Narrative: \{Narrative 2\}

....

....

\medskip
Video transcript:
\medskip

Return the most relevant narrative. If no narrative applies, return ``None''. Return only the narrative label without any additional explanation.

\end{tcolorbox}

\section{MigrationNarrate Statistics}\label{stats}
In Table~\ref{tab:dataset_size}, we present further stats about out full and labelled dataset.
\begin{table}[h]
\centering
\tiny
\begin{tabular}{lccc}
\toprule
\textbf{Dataset} & \textbf{Total Videos}  &\bf Unique Channels& \textbf{Avg. Duration (Secs)} \\
\midrule
Full corpus & \texttt{5,540}   &\texttt{2,081}&\texttt{90.46}\\
Human-annotated & \texttt{1,115} &\texttt{684} & \texttt{79.73} \\
\bottomrule
\end{tabular}
\caption{\textbf{MigrationNarrate} statistics.}
\label{tab:dataset_size}
\end{table}

\section{MigrationNarrate Examples}\label{examples_appendix}
In Table~\ref{tab:domain_examples}, we present examples from \textbf{MigrationNarrate}.

\section{Hyperparameters Fine-tuning}
\label{sec:Hyperparameters}
We fine-tuned RoBERTa-Large using standard supervised fine-tuning with a maximum sequence length of 512 tokens, batch size 16, and training for 5 epochs. We searched learning rates over $\{1e^{-5}, 2e^{-5}, 3e^{-5}\}$ and dropout values over $\{0.1, 0.2, 0.3\}$. For the LLMs, we applied QLoRA \citep{dettmers2023qlora}, where LoRA adapters were applied to all linear layers with rank $r=16$, $\alpha=32$, and dropout values from $\{0.1, 0.2, 0.3\}$. Models were trained for three epochs using learning rates from $\{1e^{-4}, 2e^{-4}, 3e^{-4}\}$, with batch size 1 and gradient accumulation of 8 steps. For all models, hyperparameters were selected based on the best macro-F1 score on the development set, and results are reported from a \textit{single run}. We present the best hyperparameters selected for each model in Table~\ref{tab:best_hyperparams}.

\begin{table}[h]
\centering
\small
\begin{tabular}{lcc}
\hline
\textbf{Model} & \textbf{Learning Rate} & \textbf{Dropout} \\
\hline
RoBERTa-large$_{\text{super\_narrative}}$  &$2e^{-5}$ & 0.3\\
RoBERTa-large$_{\text{narrative}}$  &$3e^{-5}$ & 0.1 \\
Llama-3.1-8B-Instruct &$3e^{-4}$  &0.2  \\
Qwen2.5-7B-Instruct &$3e^{-4}$  & 0.1 \\
Gemma-3-12b-it & $2e^{-4}$ & 0.1 \\
\hline
\end{tabular}
\caption{Best hyperparameter settings selected based on Macro-F1 score on the development set.}
\label{tab:best_hyperparams}
\end{table}
\section{Computational Resources}\label{gpu}
To perform our experiments, we used our internal servers
using NVIDIA A100-PCIE-
40GB GPUs.

\begin{table*}
\centering
\renewcommand{\arraystretch}{1.4}
\small
\begin{tabular}{ p{6.7cm} p{4.3cm} p{4.3cm}}
\hline
 \bf Video transcript & \bf Super-Narrative & \bf Narrative \\
\hline

PM Starmer of UK getting pressure day by day. New pressure on PM Starmer and the pressure is that one more milestone of illegals are achieved. UK is breaking records even more than US regarding illegals. 33,000 boats reached in this year. 33,000 boats mean in each boat at least 100 people may be travelling. And how many people as illegals entered? You just calculate mathematically and this is an arithmetic fashion. Illegals are being increased and Stormer feeling severe pressure from all community. Thank you. Thank you very much.
& Immigration is out of control & There are too many immigrants coming \\

\hline
 We've got a demographic crisis where the data clearly shows the number of British-born children is declining. And that's one of the reasons why lots of people are sort of saying, oh, the globalists are saying we need more mass immigration into the UK because the population... Well, maybe if you had more British-born children, then that would be a good thing. Who believe in British values, grow up with British education and understanding and culture. And that way there's less pressure to bring in immigration from overseas. So I think it makes perfect sense. And, you know, there are three key themes, family, community and country. And strong, successful working families create strong communities. And if you've got strong communities, you've got a strong country. And that's who we are and that's what we're about. I just don't care about the labels. We just want to do what we think is right for the country. 
& Immigration is part of a conspiracy & Great replacement \\

\hline
This is the situation facing not only Britain but many European cities. This is a scene from London where, for better or worse, the streets support high levels of immigrant cultures due to years of legal and illegal immigration and an unwillingness for certain cultures to integrate into traditional British culture. &Immigrants’ identity and/or culture is problematic& Certain immigrants are unwilling/incapable to integrate \\
\hline
Who are the human smugglers? In Britain, it's whoever people you are bringing in. Like if there's people coming in from Albania, them areas, it'll be Albanian gang. If they're coming from India, it's an Indian person that controls it. Right. The normal situation is 10 grand a pop and we bring you over in a container. I was talking to these people and they were saying, do you want to get involved in this, this, that, there's this money you made. They said, but what we really do is this, yeah? We get about 30 of them in a container, 10 grand a piece, we pop them off, take them to the sea, and they never get to the destination. So we don't actually do any work of taking them there. And they told you they were going to do this? This is what they do, yeah. &Immigrants are victims& Immigrants are victims of human traffickers\\
\hline
Migrant care workers in the UK are falling victim to exploitation, with some earning less than £5 an hour. Recent studies have shown that visa routes for both social and domestic workers foster conditions ripe for abuse, often leading to situations akin to modern slavery. Shocking findings include workers overstaying visas due to dire work conditions and paltry wages, with one individual on an overseas domestic worker visa receiving a mere £20 for their entire stay. Many cases reveal pay below the national minimum wage, inadequate accommodations with excessive rents, and constant surveillance. The government insists on taking action against such abusive practices. To understand the full extent of these issues and ongoing efforts, visit visaverge.com for comprehensive coverage. &Immigrants are victims&Immigrants suffer from labour exploitation\\ 

\bottomrule
\end{tabular}
\caption{Examples from \textbf{MigrationNarrate} dataset.}

\label{tab:domain_examples}
\end{table*}

\section{Annotators Consent Form}\label{consent}
We present below the questions presented in the consent form given to the annotators. The annotators were asked to answer a list of questions by either YES or NO. 
\paragraph{Taking Part in the Project}
\begin{enumerate}
    \item I have read and understood the project information sheet or the project has been fully explained to me.  (If you will answer No to this question please do not proceed with this consent form until you are fully aware of what your participation in the project will mean.)
    \item I am 18 years or older and a fluent speaker of English

   \item I have been given the opportunity to ask questions about the project. 
   \item I agree to take part in the project.  I understand that taking part in the project will include annotating content that might be triggering.

 \item I understand that I can withdraw from the study at any time; I do not have to give any reasons for why I no longer want to take part and there will be no adverse consequences if I choose to withdraw. 

\end{enumerate}
\paragraph{How my information will be used during and after the project
}
\begin{enumerate}
    \item I understand my personal details such as name and email address etc. will not be revealed to people outside the project.

    \item I understand that my personal details will not be stored, linked to my annotations or shared.
\end{enumerate}
\paragraph{So that the information you provide can be used legally by the researchers}
\begin{enumerate}
    \item I agree to assign the copyright I hold in any materials generated as part of this project to the university.
    \item I understand that the results from the annotations I provided will appear in scientific publications and white papers, without myself being identified or acknowledged, given the anonymised nature of the research.

\end{enumerate}

\begin{table}
\centering
\small
\begin{tabular}{c}
\toprule
Muslim immigration UK\\
Islam immigration UK\\
British culture immigration\\
British identity immigration\\
Sharia law immigration UK\\
immigration out of control UK\\
immigration numbers too high UK\\
mass immigration UK\\
UK losing control of immigration\\
immigration getting worse UK\\
immigrant crime UK\\
migrant crime UK\\
asylum seeker crime UK\\
immigration and terrorism UK\\
immigration national security UK\\
immigrants taking jobs UK\\
immigration jobs UK\\
asylum seekers benefits UK\\
immigrants abusing welfare UK\\
immigration burden taxpayers UK\\
immigration burden NHS UK\\
immigration housing crisis UK\\
asylum hotel cost UK\\
immigration cost of living UK\\
immigrant exploitation UK\\
immigrants suffering UK\\
immigrants discrimination UK\\
migrants victims of trafficking UK\\
migrants victims of smugglers UK\\
immigration human rights UK\\
immigration great replacement UK\\
immigration conspiracy UK\\
immigration agenda UK\\
population replacement UK\\
UK demographic replacement\\
UK needs migrant workers\\
UK needs skilled immigration\\
immigration supports the NHS\\
immigration supports UK economy\\
migrant workers essential UK\\
immigration labour shortages UK\\
immigration crackdown working UK\\
immigration policy success UK\\
immigration numbers down UK\\
border control is working UK\\
restoring border control UK\\
taking back control immigration UK\\
immigration political tool UK\\
political pressure over immigration UK\\
immigration international politics UK\\
immigration foreign policy UK\\
UK doing more on immigration\\
government ignoring immigration UK\\
politicians ignoring immigration UK\\
media hiding immigration UK\\
elites pushing immigration UK\\
sovereignty under threat immigration UK\\
immigration is the real issue UK\\
immigration remains the main issue UK
 \\
 media bias immigration UK\\
 judicial bias immigration UK\\
 system bias immigration UK\\
\bottomrule
\end{tabular}
\caption{Narrative\_oriented search phrases.}
\label{tab:narrative_search_phrases}
\end{table}
\end{document}